\documentclass[10pt,journal,compsoc]{IEEEtran}
\ifCLASSOPTIONcompsoc
  \usepackage[nocompress]{cite}
\else
  \usepackage{cite}
\fi
\usepackage{pgfplots}

\usepackage{multirow} % Required for multirows
\usepackage{booktabs} % Required for multirows
\usepackage{bbding}
\usepackage{array}
\usepackage{float}
\usepackage{makecell}
\usepackage{svg} % 需使用包
\usepackage{graphicx}
\usepackage{amsmath}

\usepackage[capitalise]{cleveref}

\usepackage{amssymb}
\usepackage{utfsym}

\usepackage{pythonhighlight}
\usepackage{algorithm}
\usepackage{algorithmic}
\ifCLASSINFOpdf
\else
\fi

\usepackage{pgf}
\begin{document}

\title{Self-supervised Training Sample Difficulty Balancing for Local Descriptor Learning}

\author{Jiahan Zhang and Dayong Tian}% <-this % stops a space
\IEEEtitleabstractindextext{%
\begin{abstract}
In the case of an imbalance between positive and negative samples, hard negative mining strategies have been shown to help models learn more subtle differences between positive and negative samples, thus improving recognition performance. However, if too strict mining strategies are promoted in the dataset, there may be a risk of introducing false negative samples. Meanwhile, the implementation of the mining strategy disrupts the difficulty distribution of samples in the real dataset, which may cause the model to over-fit these difficult samples. Therefore, in this paper, we investigate how to trade off the difficulty of the mined samples in order to obtain and exploit high-quality negative samples, and try to solve the problem in terms of both the loss function and the training strategy. The proposed balance loss provides an effective discriminant for the quality of negative samples by combining a self-supervised approach to the loss function, and uses a dynamic gradient modulation strategy to achieve finer gradient adjustment for samples of different difficulties. The proposed annealing training strategy then constrains the difficulty of the samples drawn from negative sample mining to provide data sources with different difficulty distributions for the loss function, and uses samples of decreasing difficulty to train the model. Extensive experiments show that our new descriptors outperform previous state-of-the-art descriptors for patch validation, matching, and retrieval tasks.
\end{abstract}

\begin{IEEEkeywords}
descriptor learning, gradient modulation, self-supervised learning, negative sampling unbiased processing.
\end{IEEEkeywords}}

\maketitle

% is not selected <OR> if conference mode is selected - because compsoc
% conference papers position the abstract like regular (non-compsoc)
\IEEEdisplaynontitleabstractindextext

% \fi
\IEEEpeerreviewmaketitle

% \ifCLASSOPTIONcompsoc
% % \IEEEraisesectionheading{\section{Introduction}\label{sec:introduction}}
% \else

\section{Introduction}
\label{sec:introduction}
% \fi

\IEEEPARstart{I}{n} many computer vision tasks, such as structure from motion (SfM), simultaneous localization and mapping (SLAM), image registration \cite{ma2014robust}, and 3D-reconstruction \cite{schonberger2015-3D-reconstruction}, extracting keypoints or local features from images to evaluate local correspondences is an important problem. 

To get the correspondences, there are two mainstream approaches: classical two-stage pipeline, and end-to-end pipeline. The classical two-stage pipeline consists of two steps: keypoint detection and local descriptor generation. The keypoints detection can be done by Hessian-Hessian, Difference of Gaussians (DoG), and Harris-Laplace detectors to extract keypoints. Local descriptor can be obtained by hand-crafted or learning-based methods.  End-to-end approaches have emerged in recent years, they tried to  integrate detector and descriptor as a single model \cite{yi2016lift, 2018superpoint, revaud2019r2d2, dusmanu2019D2net, 2022decoupling-Net, tyszkiewicz2020Disk}. They perform well on some benchmarks, but for reasons of computational efficiency and modular design in practice, the traditional pipeline is still competitive in the face of realistic matching scenarios \cite{jin2021frompaperToPractice}, \cite{ma2021deep-features-survey}, with the existing rich, replaceable components as well as the allowance of  incremental improvements in independent modules.
%as well as customized matchers \cite{sattler2018-6DOF-matcher, schonberger2017comparative-matcher, sarlin2020superglue} and well-established robust estimators. 

Earlier descriptors for local features were usually hand-crafted. Recently, learning-based descriptors \cite{tian2019sosnet, tian2020hynet} have been proven to be more robust than hand-crafted descriptors, especially in challenging situations with significant illumination and viewpoint changes. In practice, local descriptors learned with deep neural networks have been shown to improve the performance of two-stage pipelines \cite{ma2021deep-features-survey}. They can be seen as superior replacements for previously hand-crafted descriptors, \textit{e.g}., they can replace the SIFT feature descriptors typically used for SfM.

Overall, the goal of local descriptor learning is to sculpt a discriminative feature space in which descriptors with high matching similarity are projected to adjacent locations, while mismatched descriptors with low similarity are separated from each other. This allows us to predict whether a patch pair matches based on the distance between descriptors. Due to the limited perceptual field of image patches, the differentiation between non-matching patches and the fixed patch can be very small. For this reason, loss functions with pair-based units are usually used to improve the recognition ability of nuances in descriptor networks, such as ranking loss \cite{he2018local-ranking-loss}, n-pair loss \cite{tian2017L2Net}, and triplet loss \cite{kumar2016pair-wise-loss}. These pair-based loss functions are better suited to discriminating samples without clear distinction boundaries, and allow the model to gain the ability to discriminate small differences between matched and non-matched patches by minimizing the distance of matched/positive pairs and maximizing the distance of non-matched/negative pairs. In particular, HardNet \cite{mishchuk2017hardnet} introduced an online Hard Negative Sampling (HNS) strategy to construct harder negative pairs based on the use of triplet loss, which allowed the model to learn a more subtle gap between negative and positive examples in the dataset.

%============后面接问题吧
In subsequent studies, a number of papers followed HardNet's hard negative mining strategy and modified the design of the loss function based on a simple intuition - harder samples should receive more attention from the network to improve the model's discriminative ability further \cite{wei2018kernelized, zhang2019CDF-softmargin, tian2020hynet, wang2019EXPloss}. %Specifically, the magnitude of the loss function's weight for different pairs can be adjusted in the total loss value according to the difficulty of samples , while the difficulty can usually be measured using the relative distance between pairs. 
This idea also demonstrated its validity in the design of loss functions for other metric learning tasks \cite{50wang2017deep}, \cite{wang2018cosface}, \cite{wang2019multi-similarity-loss}.  
%============introduce problem 
However, it is worth noting that if the modulation strategy encourages too strict HNS strategy, the actual difference between the sampled positive and negative pairs may be too small, and networks that have been focusing on these extreme samples may overfit these difficult samples. This effect can be more pronounced when the model has a small amount of parameters, making it difficult for small models to learn the common paradigm used to identify samples in the dataset.
At the same time, we need to consider how to effectively measure the quality of the sampled negative samples, strong negative cases beyond a certain threshold may pick up false negatives, and these false negatives can negatively affect the model \cite{2020debiased_CL}. Using the hardest in the training dataset also shows the poor results of hardest negative mining \cite{tian2017L2Net}. To summarize, how to develop a strategy to trade off the difficulty of the extracted samples in order to provide the network learning with high-quality negative samples, is a problem worth studying.  
%Therefore, we need a more flexible modulation scheme and loss function design for HNS.
%Another problem of HNS is the forgetfulness of the model for simple samples. This is due to the fact that as the model's recognition ability increases, online HNS strategy causes the model to learn more difficult samples and fail to extract simple samples. 

We proposed XXX which tried to solve the problem from two aspects - loss function and training strategy. First, we try to introduce a dynamic-gradient-modulation strategy and use a self-supervised approach to design the loss function. This adaptive loss function combines the information generated during training, forcing the network to strike a balance between focusing on hard samples and excluding potential low-quality hard samples. %the metric depending on the different stages of its learning. 
Then, we changed the sampling strategy in the training which makes the total training process into two phases: preliminary training using the basic HNS, and annealing training (AT) afterward. we performed AT by a progressive sampling strategy and setting a threshold value to constrain the difficulty of the samples drawn from the HNS, so as to train the model with samples of progressively decreasing difficulty. In other words, the loss function aspect mainly provides fine-grained gradient modulation for the sampled data according to different learning stages, while the training strategy provides the network with data sources of different difficulty distributions. 
%Compared to most sophisticatedly designed static modulation strategies, it is more 
%and can be suitable for different datasets. This requires the network to 
%adaptively adjust the gradients during the training process according to certain metrics. 
%========改到这里:

%After the completion of preliminary training using online HNS, we perform AT by a progressive sampling strategy and setting a threshold value to constrain the difficulty of the samples drawn from the HNS, so as to train the model with samples of progressively decreasing difficulty. 
Finally, we evaluate the effectiveness of our model. The superiority of the descriptors obtained by XXX is confirmed on standard benchmarks including patch verification, matching, and retrieval tasks \cite{balntas2017Hpatches}, and the performance of our model are evaluated on downstream tasks by evaluating the pose estimation in IMC2020 \cite{jin2021frompaperToPractice}. 
The contribution of each component to the performance improvement is analyzed through ablation experiments. We also demonstrate the compatibility of the improved loss function with the training strategy, which works best when the two are combined. 
Our contributions can be summarized as follows.
\begin{itemize}
\item We proposed a balance loss for the characteristics of the data distribution of local descriptor learning, which uses dynamic gradient modulation to achieve more refined hard negative mining. %uses the information from the training stages to achieve 
\item We proposed a self-supervised strategy for sampling unbiased processing, which provides a valid discriminant for the quality of negative samples to alleviate the adverse effects of extreme values or outliers on the gradient modulation, making the model's performance improvement more significant.
\item 
We proposed a progressive sampling strategy based on difficulty to provide the network with data of different difficulty distributions. After using this annealing training strategy, the overfitting phenomenon caused by the single HNS training can be alleviated. %and the performance of the model in real scenarios can be improved.
\end{itemize}
\section{Related works}
\label{sec:related works}
\textbf{Local descriptions learning.} Early works on local patch descriptors focused on hand-crafted descriptor extraction algorithms, including SIFT \cite{lowe2004SIFT}, SURF \cite{bay2006SURF}, DAISY \cite{tian2017L2Net}, and LIOP \cite{38-2011LIOP}. With the advent of open patch datasets extracted on SIFT keypoints (i.e., Gaussian difference or DoG) \cite{brown2010Patch-dataset}, data-driven descriptor-based learning methods showed significant superiority over earlier hand-crafted methods \cite{tian2017L2Net, han2015MatchNet, mishchuk2017hardnet, tian2019sosnet}. Han \cite{han2015MatchNet} proposed MatchNet, which used a conjoined structure composed of a CNN feature network and a metric network containing three fully connected layers. The feature network was used to generate feature descriptors, and the metric network was used to learn the similarity between feature descriptors.  TFeat \cite{balntas2016TFeat} introduced triplet loss and used triplet margin loss to construct triplets. L2Net \cite{tian2017L2Net} introduced a CNN network architecture that has been widely adopted by subsequent works and redesigned the loss function and corresponding normalization. HardNet \cite{mishchuk2017hardnet} confirmed the importance of the mining strategy by using a simple but fruitful online HNS strategy to select the hardest samples from each batch to construct a triplet. SOSNet incorporated a second-order similarity measure into the loss function and combined it with the traditional first-order similarity loss term to train the descriptor network. HyNet \cite{tian2020hynet}, on the other hand, used a hybrid similarity loss that balanced the gradients from negative and positive samples, and proposed a new network architecture suitable for large-batch training. In addition to training neural networks on pre-cropped patch datasets, there are also works that exploit other cues such as global or geometric context, including ContextDesc \cite{luo2019contextdesc} and GeoDesc \cite{luo2018Geodesc}. 

\textbf{Gradient modulation.} Gradient modulation strategies are often used in metric learning to design more reasonable loss functions. Usually, in order to follow HNS strategy, the gradient of the positive pair should be modulated with an increasing function, while the gradient of the negative pair requires a decreasing function. In recent years, Circle loss \cite{sun2020circle-loss} unifies triple loss and softmax loss from a new perspective and achieves this purpose by circle margin.
For local descriptor learning, Keller \emph{et al.} \cite{keller2018learning-NHS} made each triplet axially symmetric and balance the gradients of positive and negative pairs according to the axis of symmetry. Exponential loss \cite{wang2019EXPloss} achieves the purpose of gradient modulation in exponential form so that pairs with greater relative distance receive greater attention during the update. Also, there are some related works focusing on modulation for triplet tuples according to margins. Starting from \cite{balntas2016TFeat} introducing static hard margins for local descriptor learning, Zhang and Rusinkiewicz \cite{zhang2019CDF-softmargin} further added cumulative distribution functions (CDF) to formulate dynamic soft margins. In this paper, we try to combine the modulation strategies of individual pairs and triplet tuples to make them applicable for different purposes.
\begin{figure*}[htb]
    \centering 
\includegraphics[width=.98\linewidth]{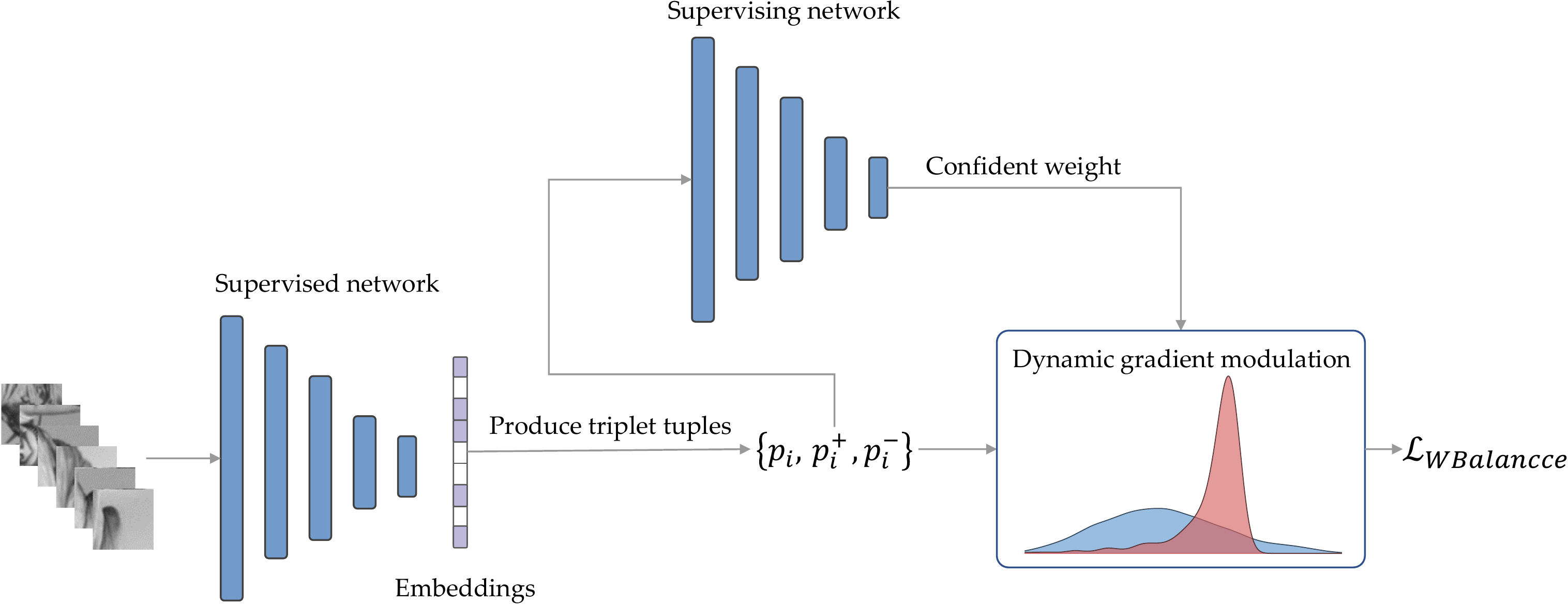}
    \caption{The calculation process of the proposed new loss function \(\mathcal{L}_{WBalance}\). }  \label{fig:preliminary training}
\end{figure*}

\textbf{Debiased Negative Sampling.} In general, when the number of positive samples and candidate negative samples is not in the same order of magnitude, the same or even better results can be achieved by using a certain strategy to sample the negative ones for selection. In other research areas, except for normal HNS, it has been shown that showing semi-hard \cite{2015facenet} or distance-weighted samples \cite{2017HardSampling} when training the model will help the improvement of performance. 
In descriptor learning, previous static negative sampling methods \cite{ST_neg_sample2012real, ST_neg_sample2012bpr} did not change with training and cannot dynamically adapt and adjust the distribution of candidate negative samples, which makes it difficult to mine more favorable negative samples. With the proposed dynamic sampling strategy, in the field of descriptor learning, Balntas \emph{et al.} \cite{balntas2016TFeat} used triple edge loss and triple distance loss for random sampling of triplet tuples. Simo-Serra \emph{et al.} \cite{simo2015discriminative}  utilized a relatively shallow architecture based on pairwise similarity to exploit hard-negative mining.  %22 23 refs in HardNet paper

The sampling strategy of HardNet \cite{mishchuk2017hardnet} outperforms classical hard-negative mining and random sampling for softmin, triplet margin, and contrastive losses. However, considering the risk of introducing false negative instances, it is important to design an effective discriminant criterion to help identify negative samples with high quality that can really improve the model performance. In our paper, this criterion is designed and incorporated into curriculum learning using a self-supervised-like approach. 
\section{Methodology}
\label{sec: Methodology}
\subsection{Method Framework}
Our novelties can be shown from two aspects - loss function and training strategy. The calculation process of our new loss function is shown in~\cref{fig:preliminary training}. For better representation, we refer to the network being trained as the supervised network and the network that provides information to guide the supervised network in training as the supervising network. The supervising network can be a descriptor network that has completed its training or can be acted by the supervised network that is currently being trained. The training patches are first input into the supervised network to generate the corresponding embeddings, then based on the distances calculated from embeddings we can select triplet tuples of batch size. 
The supervising network then recalculates the positive and negative distances of these triplet tuples and generates a confidence level for balancing the attention weight to guide the training of the supervised network. Finally, the weight produced by the supervising network will be integrated into the balance loss of the supervised network to form weighted balance loss for parameters update. 

 For the training strategy, we divide it into two stages: preliminary training and then annealing training. The main difference between them is that annealing training uses training data of decreasing difficulty, which can further improve the performance of the supervised network.
% \begin{figure}[htp] 
%     \centering
%     \includegraphics[width=4cm]{netstructure}
%     \caption{the pipeline of the preliminary training}
% \end{figure}
\subsection{Loss Function}
\begin{figure*}[htp] 
    \centering 
    \includegraphics[width=.95\linewidth]{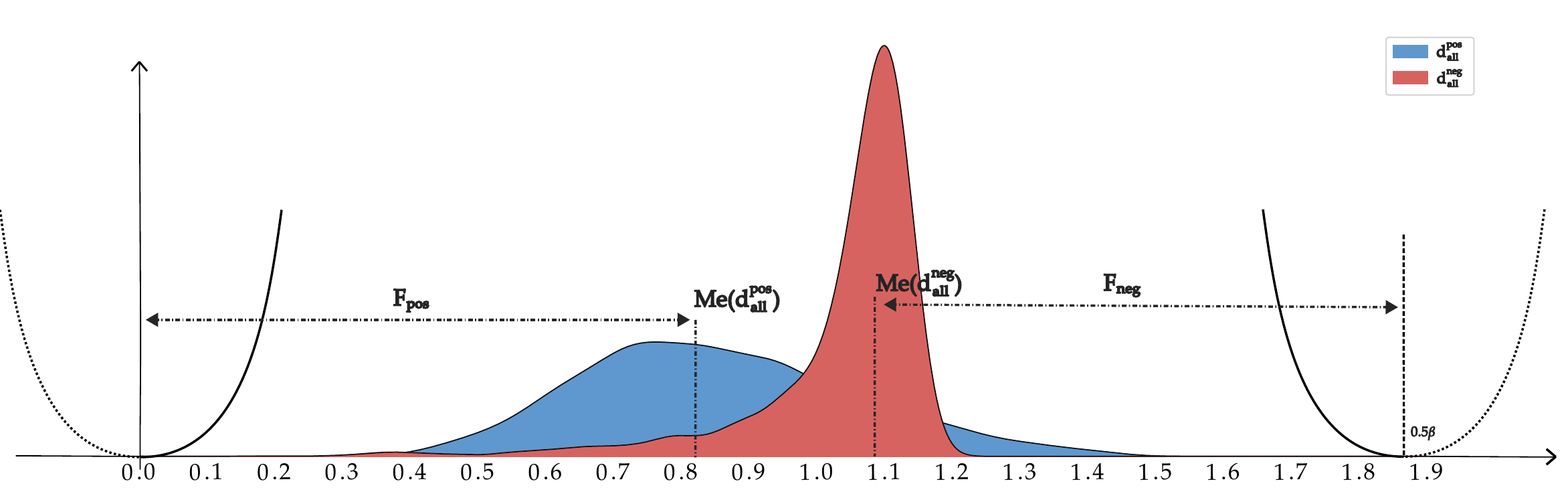}
    \caption{ Schematic of the improved $\mathcal{L}_{Balance}$ and the superimposed distance distributions of $d_{neg}$ and $d_{pos}$. The $y$-axis represents the value of the loss function or the distribution of $d_{neg}$ and $d_{pos}$ in a batch. The $x$-axis represents the \(L_{2}\) distance between the samples in a pair. The two curves represent the similarity measure functions ($S^{pos}$ and $S^{neg}$). The loss function is optimized to become smaller, which behaves like falling into two potential wells ( potential wells is a region in space where the potential energy of a particle is lower than its surrounding environment) and eventually reaching equilibrium.  } 
    \label{fig:loss overall}
\end{figure*}

Compared with triplet loss, our new loss function removes the positive margin (\textit{\textbf{t}} in~\cref{eqn:original triplet 1} from the loss function, and the strategy of gradient modulation is changed to a strategy which is similar to constructing two potential wells for positive and negative distances. 
In the process of optimizing the parameters of the supervised network along the gradient descent direction, it is like the process that the two potential wells of positive and negative distances are constantly and dynamically adjusted and finally reach relative equilibrium, as shown in~\cref{fig:loss overall}.
%d的介绍
Note that, in this paper we only use  \(L_{2}\) distance  for  descriptors similarity measure, so the negative distance and positive distance in a single triplet can be represented as:
\begin{equation} \label{L2 dist determaine}
\resizebox{0.915\linewidth}{!}{ $
\begin{array}{c}
    d_{i}^{neg}=\min{}\left(d\left({x}_{i}, {x}_{j}\right), d\left({x}_{i}, {x}_{j}^{+}\right), d\left({x}_{i}^{+}, {x}_{j}\right), d\left({x}_{i}^{+}, {x}_{j}^{+}\right)\right), \\ \\
    d_{i}^{pos}=\max{}\left(d\left({x}_{i}, {x}_{i}^{+}\right)\right), 
\end{array} $}
\end{equation}
where we have used the terms positive, negative, and anchor for corresponding \({x}_{i}^{+}\), \({x}_{i}^{+}\), \({x}_{i}\) in a triplet tuple in the following sections. Note that \({x}_{i}^{+}\) can represent more than one term if there are multiple positives, depending on how many positives are available in the dataset. When there are several positives available for an anchor in the used dataset, we will choose the positive with the largest distance from the anchor \({x}_{i}\). This is similar to a hard positive mining strategy, and we note that it can slightly improve performance.
%=====================================================
\subsubsection{Determination of similarity measure function}
\begin{figure}[htp] 
    \centering
    \includegraphics[width=.98\linewidth]{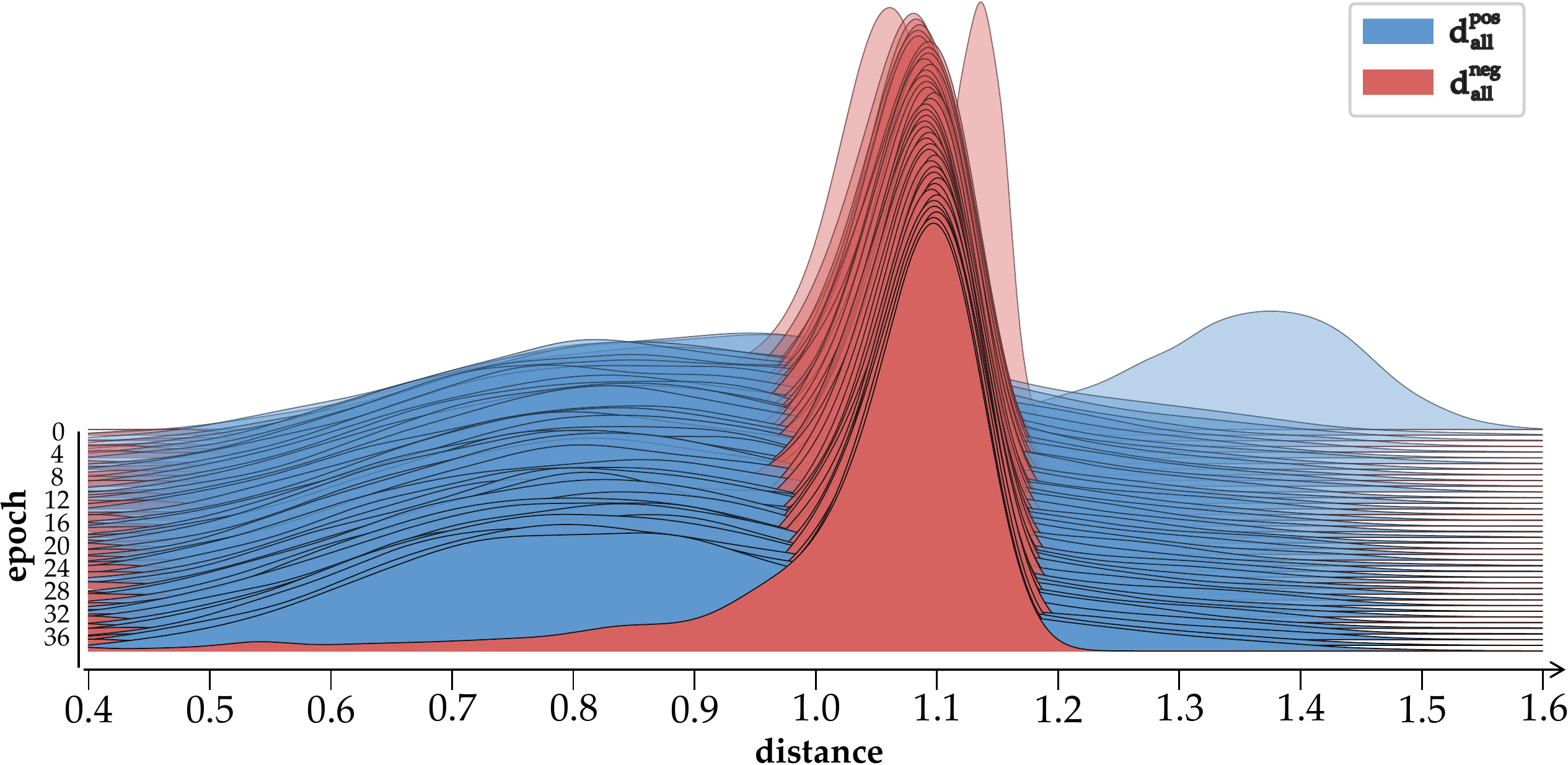}
    \caption{ Positive and negative distance distributions with different epochs. As the training progresses the positive distance distribution tends to move in the positive direction of the x-axis, while the negative distance distribution tends to move in the opposite direction.}
    \label{fig:T_all}
\end{figure}

At this stage, our overall loss function can be expressed as:
\begin{equation}
\begin{aligned}
\mathcal{L}_{Balance}=\frac{1}{N} \sum_{i=1}^{N} \left(s_{i}^{neg}+s_{i}^{neg}\right). 
\end{aligned}
\end{equation}
And the similarity measure function \(s(d)\) can be defined as two exponential functions:
\begin{equation} \label{sim measure func}
\begin{array}{c} 
s_{i}^{pos}=
\left(d_{i}^{pos} - P_{pos}\right) ^ {\alpha}, \\
\\s_{i}^{neg}=
\left(d_{i}^{neg} - P_{neg}\right) ^ {\alpha}.   \\
\end{array}
\end{equation}
The key required to characterize the similarity measure function is to determine the zero position, where is the position of the derivative of \(s(d)\) intersects the x-axis (0.5$\beta$ in~\cref{fig:loss comp}) and represented as \(P_{neg}\) and  \(P_{pos}\). As the samples of each batch change with different HNS and different model learning stages, they should be dynamically adjusted by the information from the sample distribution in a batch to make the settings of the similarity measure function at different stages more appropriate for the current model learning situation. The visualization of data distribution as training proceeds is shown in~\cref{fig:T_all}. 

According to the distribution information in a batch, We define zero positions of the similarity measure functions dynamically:
\begin{equation}
\begin{array}{c} \label{B derivation}
P_{pos} = Me \left(d_{all}^{neg}\right)-F_{pos}, \\ \\
P_{neg} = Me \left(d_{all}^{neg}\right)+F_{neg},
\end{array}
\end{equation}
where $Me$ is a method to get the median in a batch of positive or negative distances, $P_{neg}$ and $P_{pos}$ are zero positions of the corresponding similarity measure functions of negative and positive part.
$F_{pos}$ and $F_{neg}$ are focusing intensities which are used to adjust how much attention is paid to the hard samples and can be represented as distance between \(P_{pos}\) and \(Me \left(d_{all}^{neg}\right)\) shown in~\cref{fig:loss overall}. 
Based on the assumption that there is a strong correlation between these two distances which are calculated by the zero positions and the corresponding center of the distribution, we can link the two focusing intensities with:
\begin{equation} \label{eqn:ratio of 2F}
\begin{array}{c}
\gamma F_{pos} = F_{neg},
\end{array}
\end{equation}
where \(\gamma\) is a hyperparameter to adjust the ratio between the two focusing intensities. 
%In the experiments, we found that the mean and center of the distribution of positive and negative distances changed as the network was trained (shown as~\cref{fig:loss comp}). Therefore, for the consistency of the gradient modulation strategy during training, we used the dynamic statistical information from each batch to determine \({\beta}\) in each batch.  To calculate \({\beta}\), the zero position of the similarity measure function, i.e., where its derivative curve intersects the x-axis (0.5$\beta$ in~\cref{fig:loss comp}), is defined as:

For positive distribution, \({d_{i}^{neg}}\) is a positive value representing the distance between positive and anchor which should be minimized by the model. So we set $P_{pos}$ always situated at zero and only regulating \(P_{neg}\) for simplification. 
Combining~\cref{B derivation} and~\cref{eqn:ratio of 2F}, we get the formula to calculate \(P_{neg}\):
\begin{equation} \label{eqn:Pneg caculate}
\begin{array}{c}
P_{neg} = {\gamma}Me \left(d_{all}^{neg}\right) + Me \left(d_{all}^{neg}\right), \\
\end{array}
\end{equation}
where its value varies with the training process. This dynamic regulation reduces the trouble of over-parameterization and contributes to the stability of the final performance. 
%--------------------------------------------------

%========================================(记得在上文中的所有称为sample的地方改成a  triplet tuple）
\subsubsection{Adaptive weight assignment of gradient}
To make \(\mathcal{L}_{Balance}\) serve to minimize the positive distance and maximize the negative distance, we define different similarity measure functions \(s(d)\) for positive and negative distances to determine the direction and magnitude of the gradient descent, which can achieve adaptive weight assignment of gradient. The following is the proof.

\begin{figure}[htp] 
    \centering
    \includegraphics[width=.95\linewidth]{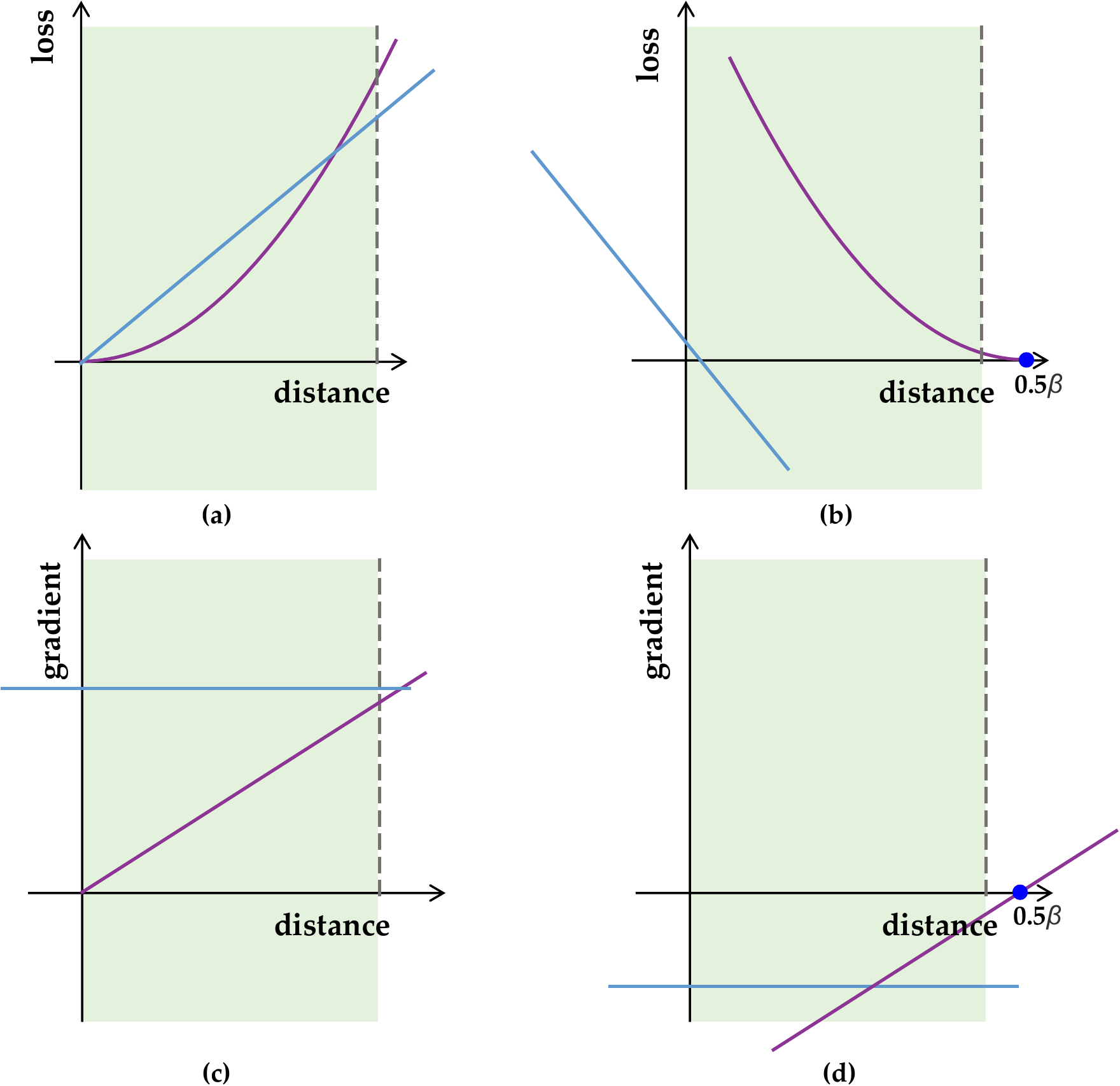}
    \caption{ Comparison of two different loss functions and their derivatives. Purple indicates our improvement, blue indicates triplet loss. The green region is the actual distribution range of \(L_2\) distance after normalization, which indicates the action range of loss functions.}
    \label{fig:loss comp}
\end{figure}

Combined with~\cref{sim measure func} and~\cref{L2 dist determaine}, the derivative of our overall loss can be represented as:
\begin{equation} \label{overall derivative}
\resizebox{0.90\linewidth}{!}{$
\begin{aligned}
\frac{\partial \mathcal{L}_{Balance}}{\partial D} & = \frac{1}{N} \sum_{i=1}^N \frac{\partial \mathcal{L}}{\partial S_i^{pos}} \left(\frac{\partial S_i^{pos}}{\partial d_i^{pos}}  \frac{\partial d_i^{pos}}{\partial p_i}+\frac{\partial S_i^{pos}}{\partial d_i^{pos}}  \frac{\partial d_i^{pos}}{\partial p_i^{+}}\right) \\
& + \frac{1}{N} \sum_{i=1}^N \frac{\partial \mathcal{L}}{\partial S_i^{neg}} \left(\frac{\partial S_i^{neg}}{\partial d_i^{neg}}  \frac{\partial d_i^{neg}}{\partial p_i}+\frac{\partial S_i^{neg}}{\partial d_i^{neg}}  \frac{\partial d_i^{neg}}{\partial p_i^{-}}\right).
\end{aligned} $}
\end{equation}
\(N\) denotes the batch size, \(D\) denotes $\left\{p_1, p_2, \ldots, p_N ; p_1^{+}, p_2^{+}, \ldots, p_N^{+}\right\}$ representing all embeddings of the corresponding patches in a batch.  \(p_i^{-}\) and \( p_i^{+}\) represent the selected patch embeddings that have the maximum or minimum distance from the anchors.
In our settings, $\frac{\partial \mathcal{L}}{\partial S_i^{pos}}$ is equal to 1 for positive and negative similarity measure functions, so the above equation can be reduced to
\begin{equation}\label{derivative_balance}
\resizebox{0.90\linewidth}{!}{ $
\begin{aligned}
\frac{\partial \mathcal{L}_{Balance}}{\partial D} & = 
\frac{1}{N} \sum_{i=1}^{N}\left(\alpha\left(d_{i}^{pos }\right)^{\alpha-1 } \frac{\partial d_{i}^{pos}}{ \partial p_{i}}+\alpha\left(d_{i}^{pos}\right)^{\alpha-1} \frac{\partial d_{i}^{pos}}{\partial p_{i}^{+}}\right) \\  
& +  \frac{1}{N} \sum_{i=1}^{N}\left( \left(\alpha\left(d_{i}^{neg} - P_{neg}\right)^{\alpha-1}\right) \frac{\partial d_{i}^{neg}}{\partial p_{i}} \right. \\
&~~~~~~~~~~~~~\left. + \left(\alpha\left(d_{i}^{neg} - P_{neg}\right)^{\alpha-1}\right) \frac{\partial d_{i}^{neg}}{\partial p_{i}^{-}}  \right).
\end{aligned} $ }
\end{equation}
Note that in~\cref{derivative_balance}, $\alpha$ is selected to range from even numbers greater than or equal to 2 and \(\alpha\left(d_{i}^{neg} - P_{ neg}\right)^{\alpha-1}\) is a negative number when it comes to actual training, so the direction of gradient decline is consistent with the triplet loss. The original triplet loss and its derivative can be represented as:
\begin{equation} \label{eqn:original triplet 1}
% \resizebox{75\linewidth}{!}{
\begin{aligned}
\mathcal{L}_{Triplet}=\frac{1}{N} \sum_{i=1}^{N} \max \left(0, t+d_{i}^{neg}-d_{i}^{neg}\right),
\end{aligned} %}
\end{equation}
\begin{equation} \label{eqn:original triplet 2}
\resizebox{0.89\linewidth}{!}{ $
\begin{aligned}
\frac{\partial \mathcal{L}}{\partial D}=\frac{1}{N} \sum_{i=1}^N\left(\frac{\partial d_i^{pos}}{\partial p_i}+\frac{\partial d_i^{pos}}{\partial p_i^{pos}}\right)
+\sum_{i=1}^N\left(-\frac{\partial d_i^{neg}}{\partial p_i}-\frac{\partial d_i^{neg}}{\partial p_i^{neg}}\right),
\end{aligned} $}
\end{equation}
when compared~\cref{overall derivative} with~\cref{eqn:original triplet 2},  $\frac{\partial S_i^{pos}}{\partial d_i^{pos}}$  and $\frac{\partial S_i^{neg}}{\partial d_i^{neg}}$  are the terms that differentiate the derivative of balance loss and the derivative of triplet loss. In \(s(d)\) the introduction of the exponential term allows the updated weight of gradient to automatically fit the size of the distance between samples. %Such an idea is similar to the concept of [ ] et al and [] et al, the difference is that we make this method more controllable to maintain the positive and negative balance.
The behaviors of triplet loss and balance loss and their derivatives are shown and compared in~\cref{fig:loss comp}.

In the backpropagation of balance loss,  due to the similarity measure functions \(s(d)\), the derivation of the network will generate additional functions with respect to the value of distances. Usually, we can set them as a primary function so that their values vary linearly with the increase or decrease of distances, as shown in~\cref{fig:loss comp} (c, d), so that the network can pay more attention and produce larger weight for embeddings with smaller \({d_{i}^{neg}}\) and larger \({d_{i}^{pos}}\) in the backpropagation process of each batch.

%=================================
\subsubsection{Negative sampling unbiased processing}
The adaptive weighting in the loss function is designed to modulate the gradient based on information about the distribution of individual pairs with positive or negative distances, which makes the network more focused on the hard negative samples. In this case, the negative sampling unbiased processing measures the relative distance in triplet pairs to mitigate the negative impact of potentially false negative samples or overly difficult negative samples so that the attention on these samples is reduced. Specifically, we propose to use a supervising network to help the model identify these outliers and reduce their weight.
%In training, we found that the performance of the model does not necessarily improve as the number of training epochs increases, and the performance of the model may even decrease after the number of triplet tuples already used for training reaches a certain number. We believe that this is due to the fact that the loss function incorrectly assigns a very large weight to the samples with outliers. By focusing on these overly difficult samples, the model may not learn anything, leading to a decrease in overall prediction performance. Therefore, we propose to use a supervising network to help the model identify these outliers and reduce their weight.

\begin{figure}[htp] 
    \centering
    \includegraphics[width=.85\linewidth]{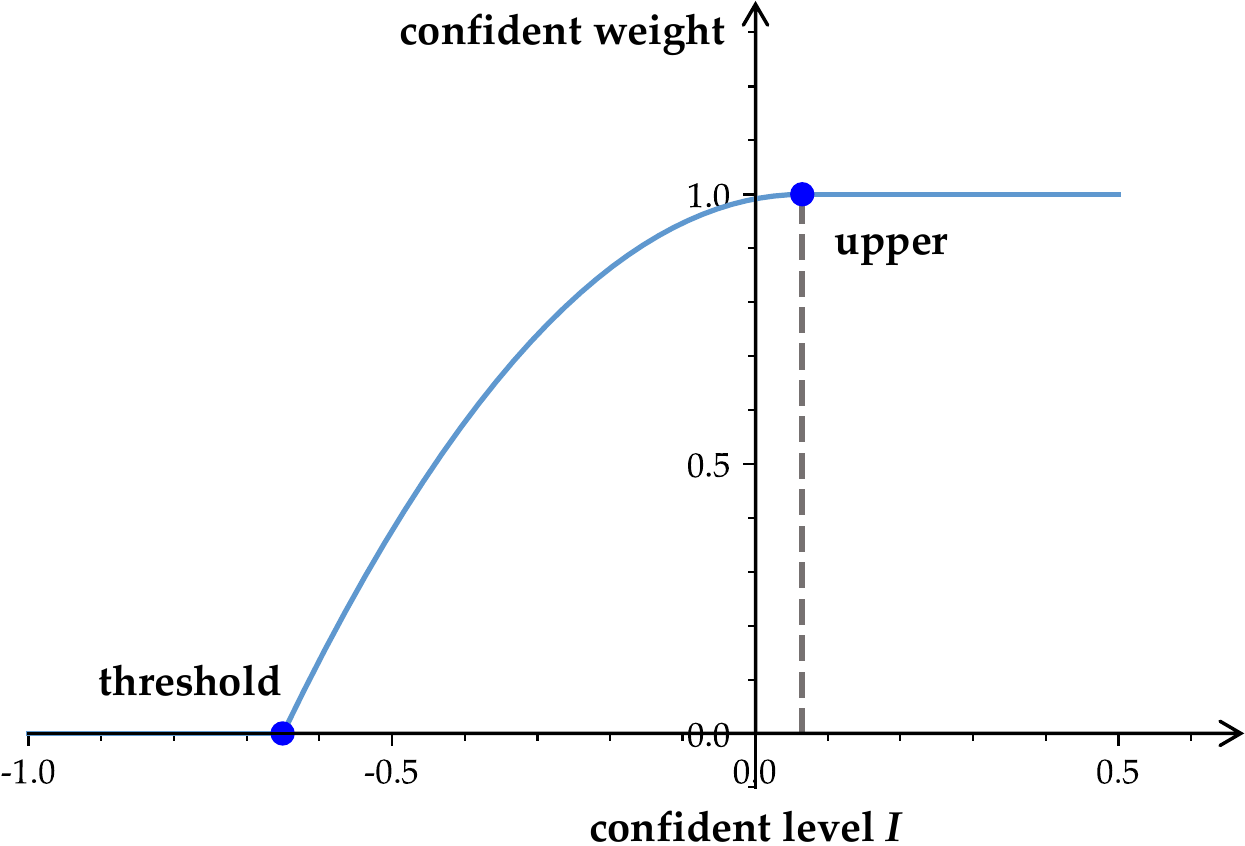}
    \caption{ The blue curve represents the values of the confident weight calculated by the \(W_c\) function with different \(I\) values.} \label{fig:mapping func}
    \label{fig:process}
\end{figure}

We can use either the training model (i.e., the supervising network and the supervised network are the same networks) or a pre-trained model, which is similar to knowledge distillation \cite{hinton2015KnowledgeDistilling}, as the supervising network. The difference is that the output of this supervising model is not labels, but the confidence level of the original labels for some samples. First, we define \(I\) to represent the confidence level of the labels for positive and negative samples, which are dynamically selected during training:
\begin{equation} \label{Eqn: I} {
I_i=d_{i}^{pos}-d_{i}^{neg}.
}
\end{equation}
Note that \({d_{i}^{neg}}\) and \({d_{i}^{neg}}\) are calculated by the embedding output of the \textbf{supervising network} for the triplet tuples which are selected by the \textbf{supervised network} through HNS.  When \({d_{i}^{neg}}\) - \({d_{i}^{neg}}\)\( < upper\), this means the anchor is predicted to be more similar to the negative than the positive ones in a triplet tuple. When the supervising network makes this judgment, we need to determine the magnitude of the weight to be mitigated for that triplet tuple based on the value of \(I\). So, we define \(W_i\) can be calculated as:
\begin{equation}
W_i=\left\{\begin{array}{cc}
f(I_i), & I_i \in[{ upper, threshold }], \\
1, & I_i > { upper}, \\
0, & I_i < { threshold },
\end{array}\right.
\end{equation}
where  \(W_i\) ranging form [0, 1] is the confident weight for each triplet tuple in a batch, and \(f(\cdot)\) is the monotonic function that maps \(I\) to the interval [0,1], which can usually be selected as an exponential function. $upper$ and $threshold$ are hyperparameters to determine the domain of the mapping function, as shown in~\cref{fig:mapping func}. %The example of actual distribution of \( W_i\) is shown in~\cref{fig:T_CW1}.

% \begin{figure}[htp] \label{fig:T_CW1}
%     \centering
%     \includegraphics[width=.95\linewidth]{T_CW1}
%     \caption{ \(W_c\) distribution in different \(I\) values.}
% \end{figure}

Finally, our weighted balance loss can be expressed as:
\begin{equation} \label{Eq:final weighted loss}
\begin{aligned}
\mathcal{L}_{WBalance}=\frac{1}{N} \sum_{i=1}^{N} \ W_i\left(s_{i}^{neg}+s_{i}^{neg}\right). 
\end{aligned}
\end{equation}
Because the supervising network only needs to process the patches selected by the supervised network and calculate the distance each time, this approach does not incur excessive computational overhead and only increases the training time by about 10\% when using the training model as the supervising network.
\subsection{Annealing training}
In the preliminary training, we used the strategy of hard negative mining and hard positive mining to select triplet tuples for gradient update. However, in the actual deployment and testing scenarios, the model needs to face a larger proportion of easy and medium difficult samples, so it is necessary to change the training strategy after the preliminary training is completed, i.e., to move from the strategy of selecting only the hardest samples to the strategy of selecting less difficult samples to improve the data generalization ability of the model in the actual testing scenarios. 

Specifically, in annealing training, the weight of these triplets will be manually set to zero when $I_i < thr$ ($I_i$ calculated from~\cref{Eqn: I}), which means that they will not participate in the parameter update of the model. This process is iterative as shown in \cref{alg: AN}. It means $bs$ and $thr$, represent the batch size and cut-off threshold, will be updated after each iteration according to:
\begin{equation}
\begin{array}{cc} \label{Eq: AN_bsUpdate_rule}
bs_{t}=bs_{t-1}-stepsize_{bs},\\
thr_{t}=thr_{t-1}+stepsize_{thr},
\end{array}
\end{equation}
until the set final value is reached. And the initial learning rate $lr$ in one iteration is calculated by:
\begin{equation} \label{Eq: initial learning rate}
\begin{array}{cc} 
lr_{t}=lr_{t-1}* \epsilon^{t},
\end{array}
\end{equation}
where $\epsilon$ is the learning rate decay factor. 
%An example of the confident weight distribution is visualized in Fig. XX. 
% Before the training we need to first set initial threshold \textbf{thr}, step size for each threshold increase \textbf{thr\_step}, end threshold \textbf{end\_thr}, batch size \textbf{bs}, step size for each batch size decrease \textbf{bs\_step}, a small learning rate \textbf{lr}, batch number of a cycle \textbf{bsNum}, learning rate decay factor \textbf{b}. The concrete implementation is as follows.
\begin{algorithm} 
	%\textsl{}\setstretch{1.8}
	\renewcommand{\algorithmicrequire}{\textbf{Input:}}
	\renewcommand{\algorithmicensure}{\textbf{Output:}}
	\caption{Annealing training algorithm}
	\label{alg1}
	\begin{algorithmic}[1] \label{alg: AN}
		\STATE Initialization: current batch size $bs$, iteration started batch size $bs_s$, iteration ended batch size $bs_e$, current iteration threshold $thr$, step size for each threshold increase $stepsize_{thr}$, step size for each batch size decrease $stepsize_{bs}$, initial learning rate $lr$
		% \STATE  ${s_{r,t}}\left( \omega  \right) = \int_0^{ + \infty } {{s_r}\left( \tau  \right){w_h}\left( {t - \tau } \right)} \exp \left( {j\omega \tau } \right)d\tau $   (via STFT)
            \STATE $n =  (bs_e - bs_s) / stepsize_{bs}$
            \STATE $t = 0 $
		\WHILE{$t < n$}
            \STATE Update $bs_t, thr_t$ based on~\cref{Eq: AN_bsUpdate_rule}
            \STATE Update $lr_t$ based on~\cref{Eq: initial learning rate}.
            \STATE Update the hyperparameters in annealing training of the current model as $bs_t, lr_t, thr_t$. 
            \STATE Start training in this iteration:
            \FOR{a triplet $i$ in a batch}
            \IF{ \( I_i < thr_t\) for triplet $i$}
            \STATE The weight of triplet $i$ will be set to zero, which means it will not participate in the process of model parameter update
            \ELSE 
            \STATE Triplet $i$ will participate normally in the process of model parameter update 
            \ENDIF
            \ENDFOR
            \STATE Construct overall loss by~\cref{Eq:final weighted loss} in a batch
            \STATE Backpropagation and model update
            % \STATE Then train the model for the set batch number
            \STATE End training in this iteration
            \STATE $t = t+1$
		% \UNTIL $n <= 0$  
            \ENDWHILE
		\ENSURE Well-trained model after annealing training
	\end{algorithmic}  
\end{algorithm}
%-------------------------------------------------
\begin{figure}[htp] 
    \centering
    \includegraphics[width=.98\linewidth]{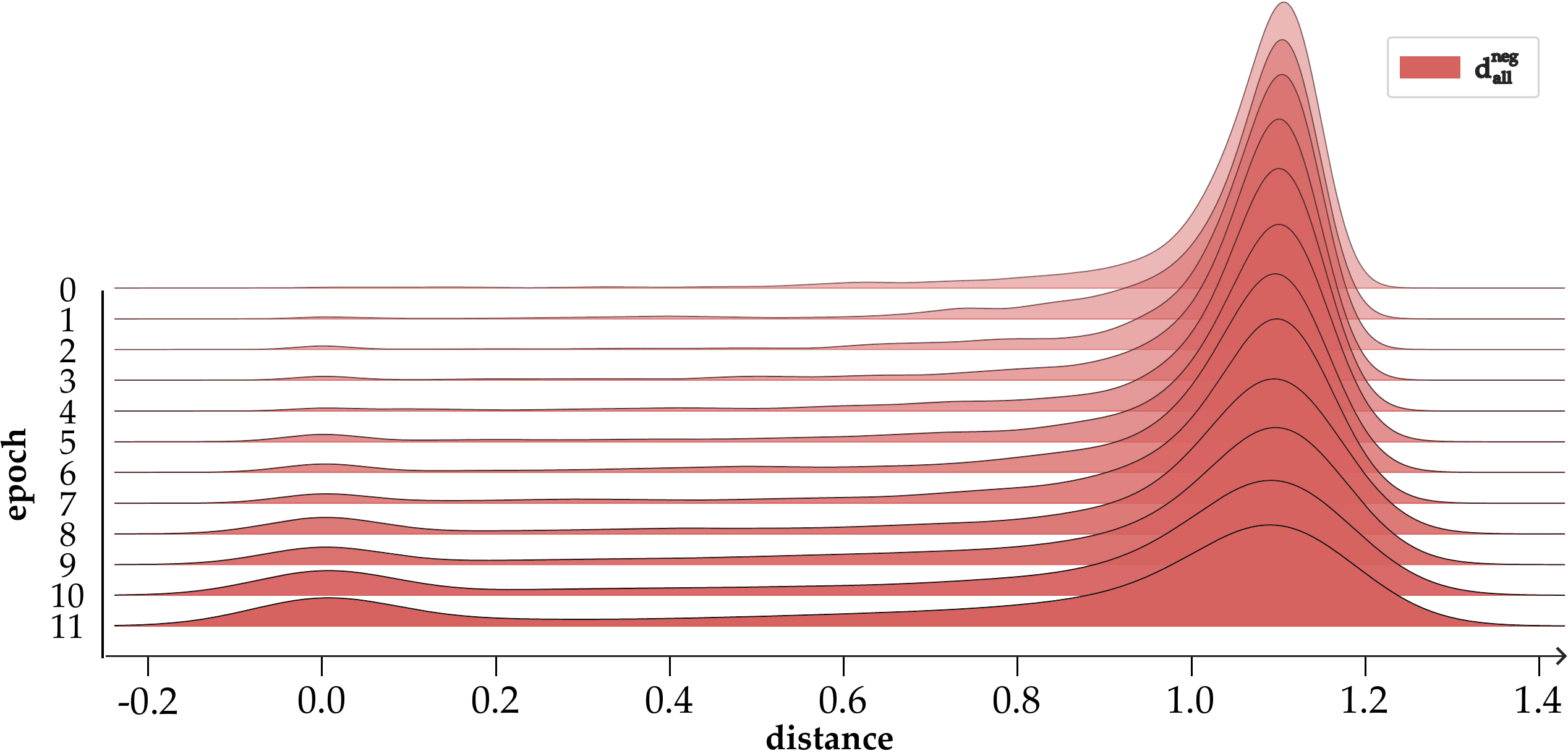}
    \caption{ Density distribution of \(W_c * {d_{all}^{neg}}\) with different epochs in annealing training. The visualization is performed by taking the first batch in each iteration. } 
    \label{fig:T_FT}
\end{figure}
%-------------------------------------------------
% \begin{python}
% '''hyperparameters: thr, end_thr, thr_step, 
% bs, bs_step, lr, bsNum, b'''
% cycle_num=(end_thr-thr)//step 
% #get the anneal cycle num
% lr_list=[lr*b for i in range(cycle_num)]
% #b is lr decay factor 
% bs_list=[bs-bs_step*(i+1) 
% thr_list=[thr+thr_step*i for i in range(cycle_num)]
% tp_list=[bsNum*bs_list[i] for i in range(cycle_num)]
% for i in (range(cycle_num)):
%   model.batch_size = bs_list[i];
%   model.tuples = tp_list[i]
%   model.threshold = thr_list[i]  
%   model.lr = lr_list[i]
%   model.train() 
% \end{python}
% \begin{python}
% def model_parameters_update():
%   # model input and output:
%   embeddings=model.generate_tuples(input_patches)
%   dist_neg,dist_pos=model.calculate_dist(embeddings)
%   confident_level=dist_pos-dist_neg
%   loss=model.loss_func(embeddings)
%   # backpropagation: 
%   for j in range(model.batch_size):
%     if confident_level[j]<model.threshold:
%    	loss[j]=0
%   model.backward(loss)
%   # if confident level<threshold, the triplet tuple
%   # will not contribute to the gradient update.
% \end{python}
The example of actual distribution of \(W_c * {d_{all}^{neg}}\) is shown in~\cref{fig:T_FT}. 
As annealing training proceeds, the samples used for training will gradually have a larger value of confidence level $I$, because the weight of triplets with low $I$ is set to 0 in \(W_c\). This is shown in the figure - the number of negative distances located at the zero position of the $x$-axis becomes larger.

The key to annealing is that the batch size gradually decreases after each iteration, while the threshold gradually increases. We define the value of \(I\) calculated from \cref{Eqn: I} for a triplet tuple as the basis for judging the difficulty of a sample. In this case, the increasing ${thr}$ will force the samples with decreasing difficulty in each batch to be used for updating the model after each iteration. Decreasing the batch size means that the probability of extracting a hard negative becomes smaller and the proportion of easy samples increases after each iteration. 
\begin{figure*}\label{fig:HPatches}
\centering
\resizebox{\textwidth}{!}{
    \includegraphics{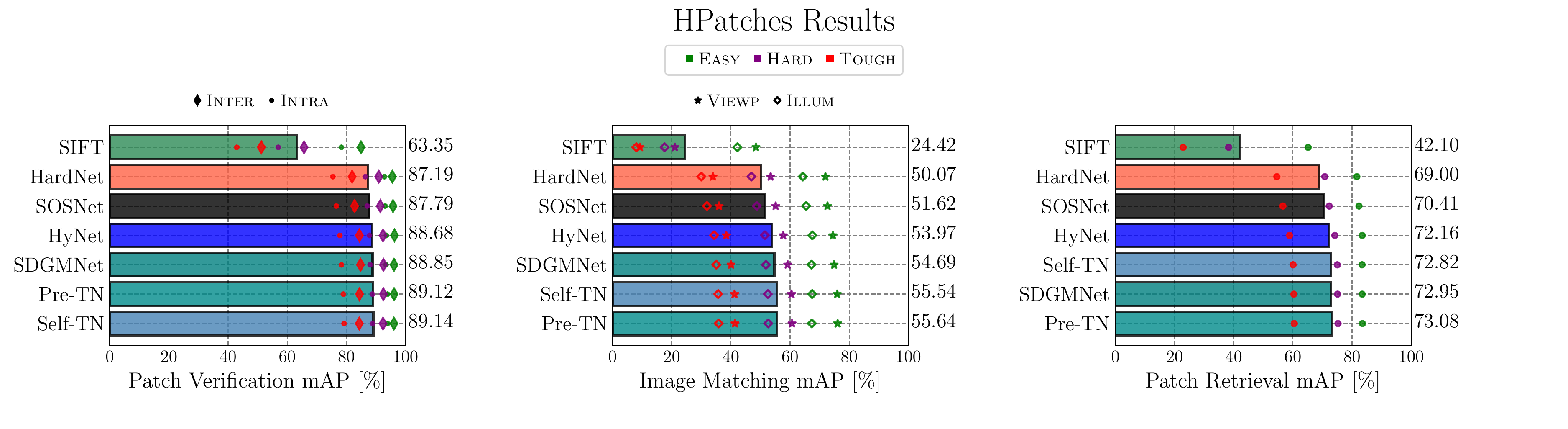}
}
 \caption{We tested the data split 'a' of the HPatches benchmark following the standard protocol. All networks were trained on the Liberty dataset from UBC PhotoTour. The bar chart shows the average scores of the networks for the three subtasks Patch Verification, Image Matching, and Patch Retrieval, evaluated as mean average precision (mAP).}
\end{figure*}

%====================================================
\section{Experiment and result}
\label{sec:Experiment}
\subsection{Training Strategy}
\begin{figure}[H]
\centering
\includegraphics[width=.98\linewidth]{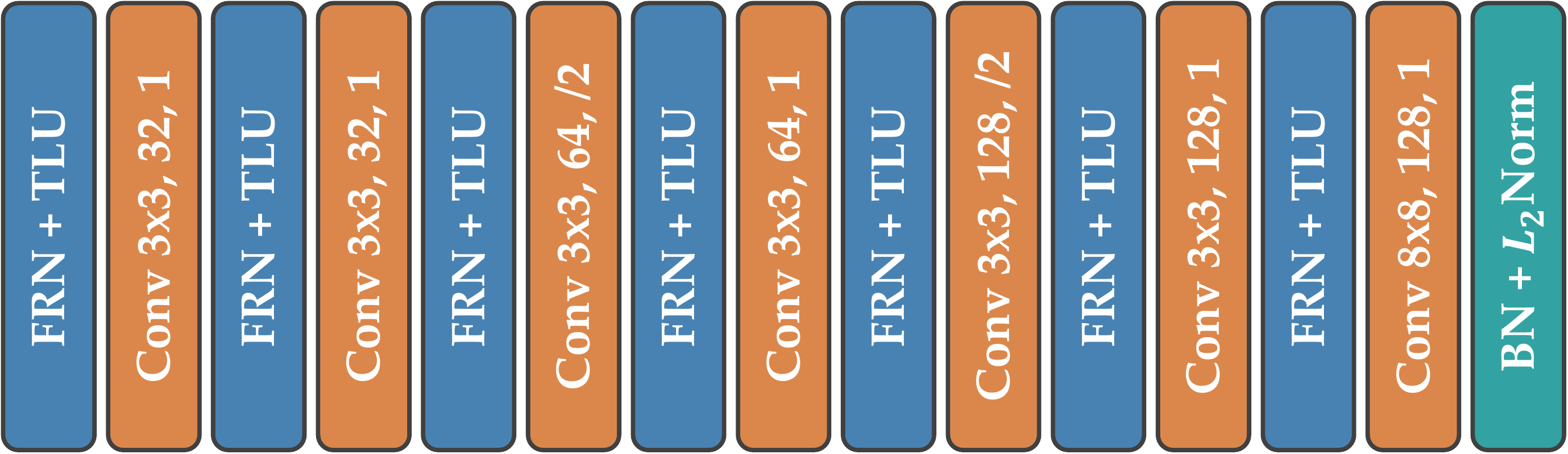}
 \caption{The network architecture is the same as that proposed by HyNet \cite{tian2020hynet} with a dropout rate of 0.3.} 
 \label{fig:Arch_network}
\end{figure}
\textbf{Preliminary training}. We used PyTorch to implement XXNet. As shown in~\cref{fig:Arch_network}, for a fair comparison, the network architecture is the same as that proposed by HyNet \cite{tian2020hynet} and has the same amount of parameters as HardNet \cite{mishchuk2017hardnet}. In the specific implementation of balance loss, we set $\alpha$=2, $\gamma$=1.05, and in the process of outliers we set supervising network with $upper$=0.065 and $threshold$=-0.65. The optimal model was trained with 25 epochs with 4050000 tuples per epoch and batch size=2560. Adam optimizer was used to update parameters with max lr=0.033, specifically using linear warm-up and cosine decay strategy for learning rate.

In~\cref{sim measure func}, \(\alpha\) is a scalar. Normally, \(\alpha\) can be set equal to larger than 1 to represent an exponential function. We found a number larger than 2 does not necessarily lead to better results, so in the following experiments, we always set \(\alpha\) equal to 2. 

\textbf{Annealing training.} In AT, we tried to use triplet tuples of decreasing difficulty to update the network parameters at small learning rates. Specifically, the decrease in difficulty was achieved by setting the batch size to decrease by $stepsize_{bs}$=128 per iteration and the threshold to increase by $stepsize_{thr}$=0.05. Other parameters included: $bs_s$=2944, $bs_e$=1024, initial $thr$=-0.05, number of batches per iteration=1400. We also set the initial $lr$=\(1.5*10^{-6}\) with a decay factor $\epsilon$=0.75 for each iteration.

\subsection{Experiment Settings}
We compared the performance of our solution with the state of the art on three benchmarks: UBC benchmark \cite{brown2010Patch-dataset}, HPatches \cite{balntas2017Hpatches} and IMC2020 \cite{jin2021frompaperToPractice}. We implemented two versions of the scheme, one using the training model as the supervising network called Self-TN and the other using a pre-trained supervising network with more parameters called Pre-TN.
% \subsubsection{Datasets}

\textbf{UBC Benchmark} evaluated on the Brown dataset \cite{brown2010Patch-dataset}. It is a widely used patch dataset for evaluating the performance of local descriptors. The dataset consists of three subsets for three different scenarios: Liberty, Yosemite, and Notredame. Typically, deep descriptor networks are trained on one subset and tested on the other two subsets. According to the standard protocol \cite{brown2010Patch-dataset}, the test verifies that the network can correctly determine the 100K matching and non-matching pairs in the two test subsets. The protocol evaluates the metric using the false-positive rate at 95\% recall (FPR@95), and we reported the experimental results for our two versions, including the average over all subsets, in~\cref{tab: UBC dataset}. 

\begin{table*}[htbp]
	\centering
	\renewcommand\arraystretch{1.2}
	%\setlength{\abovecaptionskip}{0.cm}
	%\setlength{\belowcaptionskip}{-0.cm}
	% \textbf{Table 1}\\~~PATCH VERIFICATION PERFORMANCE ON UBC PHOTOTOUR. NUMBERS SHOWN ARE FPR@95 ($\%$) THAT ARE LOWER FOR BETTER. THE BEST SCORES ARE HIGHLIGHTED IN BOLD. DASH LINES INDICATE CHANGES OF MODELS. LIB: Liberty, YOS: Yosemite, ND: Notredame.\\  
            \caption{~~PATCH VERIFICATION PERFORMANCE ON UBC PHOTOTOUR. NUMBERS SHOWN ARE FPR@95 ($\%$) THAT ARE LOWER FOR BETTER. THE BEST SCORES ARE HIGHLIGHTED IN BOLD. DASH LINES INDICATE CHANGES OF MODELS. LIB: Liberty, YOS: Yosemite, ND: Notredame.}  
            \label{tab: UBC dataset} %%表的标题

    %\resizebox{\linewidth}{!}{    % Resize table to fit within
            % \documentclass{article}

% %...\usepackage{amsmath}
% \usepackage{multirow} % Required for multirows
% \usepackage{booktabs} % Required for multirows
% %\usepackage{multicol}
% \usepackage{array}
% \begin{document}
	
% 	%...
	
% \begin{table}[hp] %%参数： h:放在此处 t:放在顶端 b:放在底端 p:在本页
	% \centering
	% \renewcommand\arraystretch{1.2}
	% %\setlength{\abovecaptionskip}{0.cm}
	% %\setlength{\belowcaptionskip}{-0.cm}
	%   % 显示位置为中间	
	% \label{TABLE 1}  % 用于索引表格的标签
	% %\caption{	}  % 表格标题
	% \textbf{Table 1}\\~~PATCH VERIFICATION PERFORMANCE ON UBC PHOTOTOUR. NUMBERS SHOWN ARE FPR@95 ($\%$) THAT ARE LOWER FOR BETTER. THE BEST SCORES
	% ARE HIGHLIGHTED IN BOLD. DASH LINES INDICATE CHANGES OF MODELS. LIB: Liberty, YOS: Yosemite, ND: Notredame.\\  %%表的标题
	% %字母的个数对应列数，|代表分割线
	% % l代表左对齐，c代表居中，r代表右对齐
	% %%表的标题
	
	\begin{tabular}{>{\centering}p{70pt}|cccccccc|c} %第一列设置宽度为45pt 全为左对齐 没有分割线
		\hline  % 表格的横线
		%\toprule % 顶部线
		Train & ND & YOS & & LIB & YOS & & LIB & ND & \multirow{2}{*}{\textbf{Mean}}  \\%[3pt]只改一行    %%表格第一行标题 % 表格中的内容，用&分开，\\表示下一行
		%\hline  % 表格的横线
		\cline{1-2}\cline{2-3}\cline{5-6}\cline{8-9}
		Test & \multicolumn{2}{c}{LIB} & & \multicolumn{2}{c}{ND} & & \multicolumn{2}{c|}{YOS} & \multirow{2}{*}{}\\
		\hline
		%\cmidrule(r){3-4}  \cmidrule(r){5-6} \cmidrule(r){7-8}
		SIFT \cite{lowe2004SIFT} & \multicolumn{2}{c}{29.84} & & \multicolumn{2}{c}{22.53} &  &    \multicolumn{2}{c|}{22.79} & 26.55\\
		\hline
		TFeat \cite{balntas2016TFeat} & 7.39 & 10.13 & & 3.06 & 3.80 & &  8.06 & 7.24 & 6.64  \\
		\hline
		L2Net \cite{tian2017L2Net} & 2.36 & 4.70 & & 0.72 & 1.29 & & 2.57 & 1.17 & 2.23  \\
		\hline
		HardNet \cite{mishchuk2017hardnet} & 1.49 & 2.51 & & 0.53 & 0.78 & & 1.96 & 1.84 & 1.51  \\
		\hline
		CDFDesc \cite{zhang2019CDF-softmargin} & 1.21 & 2.01 & & 0.39 & 0.68 & & 1.51 & 1.29 & 1.38  \\
		\hline
		SOSNet \cite{tian2019sosnet} & 1.08 & 2.12 & & 0.34 & 0.67 & & 1.03 & 0.95 & 1.03  \\
		\hline
	    % HN$-$FRN[28][46] & 1.26 & 1.76 & & 0.41 & 0.58 & & 1.16 & 1.05 & 1.04  \\
	    % \hline
	    HyNet \cite{tian2020hynet} & 0.89 & 1.37 & & 0.34 & 0.61 & & 0.88 & 0.96 & 0.84  \\
	    \hline
	    Ours & 0.86 & 1.32 & & 0.30 & 0.57 & & 0.80 & 0.91 & 0.79  \\
	    \hline
		%\bottomrule % 底部线
		%\hline  % 表格的横线
	\end{tabular}
% \end{table}
% \end{document}

            % \floatfoot{\textit{Note}: Table notes.}
    %}
\end{table*}
% \textbf{AMOS Dataset}: In the first phase, we use images from a variety of different lighting conditions for training, including the AMOS [ ] dataset, and our own dataset, which is a large set of image clippings containing patches of illumination and appearance variations, mainly for training robust local feature descriptors. consists of 26 sets of data for training and 7 sets of images for testing, each set consisting of 50 patches under different lighting conditions

\textbf{HPatches} \cite{balntas2017Hpatches} is a more comprehensive benchmark that evaluates descriptors on three tasks: patch verification, image matching and patch retrieval. According to geometric distortion, subtasks are categorized into Easy, Hard and Tough. Furthermore, patch pairs from the same or different image sequences are separated into two test subsets for verification, denoted by Intra and Inter, respectively. And the matching task is designed to evaluate the viewpoint (VIEWP) and illumination (ILLUM) invariance of descriptors. 

\textbf{Image Matching Challenge (IMC).} IMC2020 is a large-scale challenge dataset of wide baseline matching \cite{jin2021frompaperToPractice}. IMC2020 consists of datasets with large illumination and viewpoint variations. In this experiment, we compare our method with the existing descriptors in an two-stage image matching pipeline. We experiment on the stereo track using the validation sets of Phototourism. This benchmark takes the predicted matches as input and measures the 6 DoF pose estimation accuracy. We measure the mean average accuracy (mAA) of pose estimation at $10^{\circ}$ and the number of inliers. 
% \begin{table}[H]  %%参数： h:放在此处 t:放在顶端 b:放在底端 p:在本页
% 		\centering
% 		\renewcommand\arraystretch{1.2}
% 		%\setlength{\abovecaptionskip}{0.cm}
% 		%\setlength{\belowcaptionskip}{-0.cm}
% 		% 显示位置为中间	
%               % \textbf{Table 2}\\~~IMC2020 performance on validation set\\ 
% 		\caption{Performance (mAA) on the validation set of IMC2020.}   % 表格标题 
%             \label{tab:IMC_result} % 用于索引表格的标签
% 		%\caption{}  % 表格标题
% 		% \textbf{Table 1}\\~~PATCH \\  %%表的标题
% 		%字母的个数对应列数，|代表分割线
% 		% l代表左对齐，c代表居中，r代表右对齐
% 		%%表的标题
%         \resizebox{\linewidth}{!}{    % Resize table to fit within
% 		\begin{tabular}{>{\centering}p{70pt}|cc|cc|c} %第一列设置宽度为45pt 全为左对齐 没有分割线
% 			%\setlength{\tabcolsep}{20mm}
% 			\hline  % 表格的横线
% 			%\toprule % 顶部线
% 			\#Keypoints &  \multicolumn{2}{c|}{\textbf{2k}} & \multicolumn{2} {c|}{\textbf{8k}}& \multirow{2}{*}{\textbf{Mean}}  \\%[3pt]只改一行    %%表格第一行标题 % 表格中的内容，用&分开，\\表示下一行
% 			%\hline  % 表格的横线	
% 			\cline{1-5}
% 			Task & Stereo & Mutiview & Stereo  & Mutiview & \multirow{2}{*}{}\\
% 		    \hline  % 表格的横线
% 		    HardNet&54.32&61.64&67.07&74.24&64.32\\
% 		    \hline
% 		    SosNet&54.33&61.11&66.97&74.56&64.24\\
% 		    \hline
% 		    HyNet&54.71&62.13&67.29&74.78&64.73\\
% 		    \hline
% 		    Ours&55.62&63.51&68.13&75.53&65.70\\
% 		    \hline
% 		\end{tabular}
%   }
% \end{table}
\section{Discussion}
%-------------------------------------------看到：
In this section, we discuss the contribution of each component of XXNet to the overall performance and its influencing factors. Specifically, we trained different models on Liberty \cite{brown2010Patch-dataset} and tested them on the Hpatches \cite{balntas2017Hpatches} matching task to obtain the influence of our three modifications - loss function, negative sampling unbiased processing, and annealing training - on the performance of the model.

Furthermore, we observed that a better FRP@95 metric does not necessarily imply a better mAP metric, which we believe is due to the fact that the FRP@95 metric is more concerned with the model's ability to identify difficult samples and does not reflect how well the model fits all samples. This is related to the way FRP@95 is calculated. To better assess the performance of the model on all samples, we use mAP as the metric for the following test.

\subsection{Ablation Study}          	
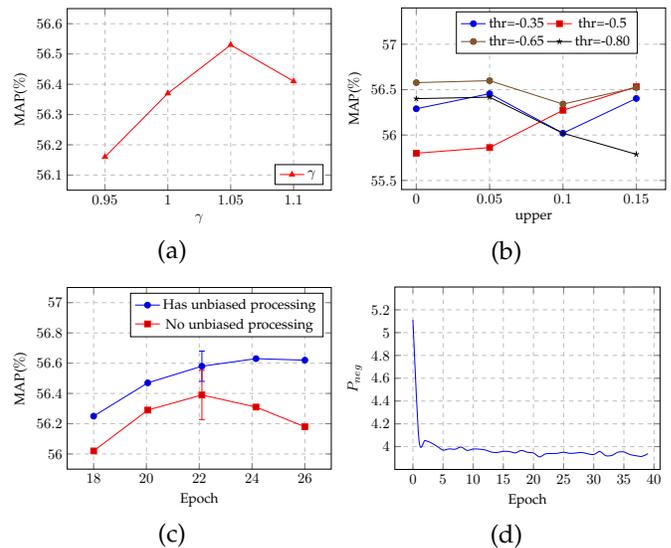
\begin{figure}
    \centering
    \begin{minipage}[c]{0.5\linewidth}
    \centering
%%%%%%%%%%%%%%%%%%%%%%%%%%%
% \subfigure{
  %   	\begin{figure}	\label{fig:param_gamma}
		% \centering
  \resizebox{\linewidth}{!}{
    % \documentclass[a4paper]{article}

% for dvipdfm:
%\def\pgfsysdriver{pgfsys-dvipdfm.def}
% \usepackage{pgfplots}
% \usepackage{subfigure,graphicx}
% \usepackage{tikz}
\pgfplotsset{width=8cm,height=6cm,compat=1.5}% <-- moves axis labels near ticklabels (respects tick label widths)
% \usepackage{amsmath}
% \usepackage{float}
% \usepackage[figuresright]{rotating}
% \usepackage[margin=1in]{geometry}
% \usepackage[justification=centering]{caption}
% \pgfplotsset{compat=newest}
% \usetikzlibrary{plotmarks}
% \usetikzlibrary{arrows.meta}
% \usepgfplotslibrary{patchplots}

% \begin{document}

%     wdadwad

	% \begin{figure}	\label{fig:param_gamma}
		\centering
		\begin{tikzpicture} 
			\begin{axis}[
				title={},
				xlabel={${\gamma}$},
				ylabel={MAP($\%$)},
				xmin=0.92, xmax=1.13,
				ymin=56.05, ymax=56.65,
				xtick={0.95, 1.0, 1.05, 1.10},
				ytick={56.1, 56.2, 56.3, 56.4, 56.5, 56.6},%TODO:让具体点的值显示在图上，还有这个图的位置不太对
				legend pos=south east,
				xmajorgrids=true,
				ymajorgrids=true,
				grid style=dashed,
				]
				\addplot[mark=triangle*,red]
				coordinates {
					%(0,51.9)
					(0.95, 56.16)
					(1.0, 56.37)
					(1.05, 56.53)
					(1.10, 56.41)
				};
				\legend{${\gamma}$}
			\end{axis}
			
		\end{tikzpicture}
		% \caption{Trend of MAP(\%) with different ${\gamma}$ values}
	% \end{figure}

% \end{document}
  }
  	{(a)}
		% \caption{Trend of MAP(\%) with different ${\gamma}$ values}
	% \end{figure}
%%%%%%%%%%%%%%%%%%%%%%%%%%%
    \end{minipage}%
    % \hspace{10mm}  % 用于控制间距
    \begin{minipage}[c]{0.5\linewidth}
    \centering
%%%%%%%%%%%%%%%%%%%%%%%%%%%
    % \begin{figure}	
    % \centering
% \subfigure{
  \resizebox{\linewidth}{!}{
    \pgfplotsset{width=8cm,height=6cm,compat=1.5}% <-- moves axis labels near ticklabels (respects tick label widths)

    \begin{tikzpicture}
			\begin{axis}[
				title={},
				xlabel={upper},
				ylabel={MAP($\%$)},
				xmin=-0.01, xmax=0.17,
				ymin=55.4, ymax=57.4,
				xtick={ 0, 0.05, 0.10, 0.15},
                xticklabels={$0$,$0.05$,$0.1$,$0.15$},
				ytick={55.5, 56.0, 56.5, 57.0},
                    legend style={at={(0.55,0.98)},
	            anchor=north, legend columns=2},
				xmajorgrids=true,
				ymajorgrids=true,
				grid style=dashed,
				]
			 % \addplot
% 				coordinates {
% % 					(0,51.9)
% % 					(1,53.5)
% % 					(2,54.1)
% % 					(3,53)
% % 					(4,52.5)
% 				};
        	\addplot
				coordinates {
					(0, 56.2895)
					(0.05, 56.4555)
					(0.1, 56.0203)
					(0.15, 56.4031)
				};
				\addplot        %thr=-0.5
				coordinates {
 					(0,55.8)
 					(0.05,55.86215)
 					(0.1,56.2727)
					(0.15,56.5319)
				};

				\addplot        % thr=-0.65
				coordinates {
					(0, 56.5772)
					(0.05,56.5987)
					(0.1,56.3412)
					(0.15,56.5227)
				};
                \addplot
				coordinates {
					(0, 56.401)
					(0.05, 56.4171)
					(0.1, 56.0203)
					(0.15, 55.7882)
				};
				\legend{thr=-0.35, thr=-0.5,  thr=-0.65, thr=-0.80}
			\end{axis}
		\end{tikzpicture}
}
	{(b)}
% 		\caption{Trend of MAP(\%) with different thr and upper values}
	% \end{figure}
%%%%%%%%%%%%%%%%%%%%%%%%%%%
    \end{minipage}%
    \vspace{3mm}
    \begin{minipage}[c]{0.5\linewidth}
    \centering
%%%%%%%%%%%%%%%%%%%%%%%%%%%
    % \begin{figure}
    % \centering
% \subfigure{
\resizebox{\linewidth}{!}{
    \pgfplotsset{width=8cm,height=6cm,compat=1.5}% <-- moves axis labels near ticklabels (respects tick label widths)

\begin{tikzpicture}
        \begin{axis}[
        title={},
		xlabel={Epoch},
		ylabel={MAP($\%$)},
		xmin=17, xmax=27,
		ymin=55.9, ymax=57.1,
			xtick={ 18, 20,22,24,26},
			ytick={56.0,56.2,56.4,56.6,56.8,57.0},
			legend pos=north east,
			xmajorgrids=true,
			ymajorgrids=true,
			grid style=dashed,
			]
			
     	\addplot+ [
    	error bars/.cd,
	    y dir=both, y explicit
	    ]
	     coordinates {
	     	(18,56.25)  
	    	(20.05,56.47)  
	    	(22.1,56.58)  +-(0.1,0.1001)
	    	(24.15,56.63)
	    	(26,56.62)  
    	};
	    
	    	\addplot+ [
	    	error bars/.cd,
	    	y dir=both, y explicit
	    	]
	    	coordinates {
	    		(18,56.02)  
	    		(20.05,56.29)  
	    		(22.1,56.39)  +-(0.1,0.1626)
	    		(24.15,56.31)
	    		(26,56.18)    
	    	};
		
				\legend{Has unbiased processing, \makecell[c]{No unbiased processing}}
			\end{axis}
		\end{tikzpicture}
}
	{(c)}
% \end{figure}
%%%%%%%%%%%%%%%%%%%%%%%%%%%
    \end{minipage}%
    % \hspace{10mm}  % 用于控制间距
    \begin{minipage}[c]{0.5\linewidth}
    \centering
%%%%%%%%%%%%%%%%%%%%%%%%%%%
% \begin{figure}  \scriptsize
	%\begin{figure}
	% \centering
 % \subfigure{
\resizebox{\linewidth}{!}{
    \pgfplotsset{width=8cm,height=6cm,compat=1.5}% <-- moves axis labels near ticklabels (respects tick label widths)

 \begin{tikzpicture}
		\begin{axis}[
			title={},
			xlabel={Epoch },
			ylabel={\(P_{neg}\) },
			xmin=-3, xmax=41,
			ymin=3.8, ymax=5.4,
			xtick={0,5,10,15,20,25,30,35,40},
			ytick={4.0,4.2,4.4,4.6,4.8,5.0,5.2},
			legend pos=north east,
			xmajorgrids=true,
			ymajorgrids=true,
			grid style=dashed,
			]
			\addplot+[mark= ,smooth]
			coordinates {
				(0,	5.111057281)
				(1	,4.05937767)
				(2	,4.053604603)
				(3	,4.03431654)
				(4	,4.002567291)
				(5	,3.970087051)
				(6	,3.979952335)
				(7	,3.977294207)
				(8	,3.996511459)
				(9	,3.967026472)
				(10	,3.979092121)
				(11,3.977762699)
				(12	,3.971040726)
				(13	,3.95240736)
				(14	,3.948887587)
				(15	,3.958926439)
				(16	,3.956604958)
				(17	,3.945422649)
				(18	,3.966887474)
				(19	,3.950642109)
				(20	,3.944139481)
				(21	,3.91023159)
				(22	,3.936540604)
				(23	,3.939357519)
				(24	,3.941304684)
				(25	,3.950718403)
				(26	,3.940989017)
				(27	,3.943984032)
				(28	,3.949210167)
				(29	,3.9370718)
				(30	,3.931236506)
				(31	,3.957084656)
				(32	,3.920632362)
				(33	,3.92252779)
				(34	,3.951244831)
				(35	,3.953077316)
				(36	,3.930819988)
				(37	,3.920306206)
				(38	,3.914366245)
				(39	,3.938398838)
			};
			%\legend{m=0.9,m=1.0,m=1.1}
		\end{axis}		
	\end{tikzpicture}
 }
	{(d)}
	% \centering
	%\end{figure}
% \end{figure}
%%%%%%%%%%%%%%%%%%%%%%%%%%%
    \end{minipage}
	\caption{Experiments on the effect of hyperparameters on performance after preliminary training. The four figures are (a): The effect of hyperparameter $\gamma$ on the model performance. (b): The effect of hyperparameters $threshold$ and $upper$ on the model performance. (c): The effect of adding the supervising network (TNet) on the balance loss. (d): Adaptive changes of $P_{neg}$ values during gradient modulation.}
    \label{fig:4ABL-study-fig}
\end{figure}

	\begin{table}[H] %%参数： h:放在此处 t:放在顶端 b:放在底端 p:在本页
		\centering
		\renewcommand\arraystretch{1.2}
		%\setlength{\abovecaptionskip}{0.cm}
		%\setlength{\belowcaptionskip}{-0.cm}
		% 显示位置为中间	
		% \textbf{Table 1} \\  %%表的标题
		\caption{Ablation experiments to explore the performance gains of each modification. The result is tested on data split 'full' from the HPatches benchmark. mAP score for subtask image matching is reported.}  % 表格标题
            \label{tabl:Abl}  % 用于索引表格的标签
		\begin{tabular}{>{\centering}p{60pt}|c|c|c} %第一列设置宽度为45pt 全为左对齐 没有分割线
			%\setlength{\tabcolsep}{20mm}
			%\hline  % 表格的横线
			%\toprule % 顶部线
			Modification & Choice & \makecell[c]{Add unbiased\\processing} & MAP(\%) \\%[3pt]只改一行    %%表格第一行标题 % 表格中的内容，用&分开，\\表示下一行
			\hline  % 表格的横线
			
			\multirow{2}{*}{} & Triplet loss & $\usym{2717}$ &55.67\\
                \multirow{2}{*}{\textbf{Loss function}} & Triplet loss & $\checkmark$ &56.08\\
			\multirow{2}{*}{} & \makecell[c]{Balance loss} & $\usym{2717}$ & 56.53 \\
                \multirow{2}{*}{} & Balance loss & $\checkmark$ & 56.83\\
			\hline
			
			% \multirow{2}{*}{\textbf{{\makecell[c]{Unbiased processing\\ with TNet}}}} & %Triplet loss+TNet &  56.08\\
			% \multirow{2}{*}{} & Balance loss+TNet & 56.83\\
			% \hline
			
			\multirow{3}{*}{\textbf{{\makecell[c]{Annealing\\training}}}} & AT+Triplet loss & $\usym{2717}$ & 56.25\\
			\multirow{3}{*}{} & AT+Triplet loss & $\checkmark$ & 56.59\\
			\multirow{3}{*}{} & \makecell[c]{AT+Balance loss} & $\usym{2717}$ &57.01\\
			\multirow{3}{*}{} & AT+Balance loss & $\checkmark$ & 57.35\\
			\hline
			
% 			\multirow{2}{*}{\textbf{{\makecell[c]{Descriptor\\type}}}} & {\multirow{2}*{\makecell[c]{Hard+FRN\\SOSNet+FRN}}} &\\
% 			\multirow{2}{*}{} & {\multirow{2}*{}}& \\
% 			\hline
			
			%\bottomrule % 底部线
			%\hline  % 表格的横线
		\end{tabular}
	\end{table}
\textbf{Loss function. } The performance improvement of balance loss compared with triplet loss is shown in~\cref{{tabl:Abl}}, and it is more significant after adding unbiased processing. Also, unbiased processing can be applied to the original triplet loss to improve performance. The influence of the magnitude of $\gamma$ value on the final mAP performance of the model can be seen in the~\cref{fig:4ABL-study-fig}(a). 
According to the previous~\cref{eqn:Pneg caculate}, the $\gamma$ can be interpreted as a parameter here that balances the update weight between positive and negative distances. Specifically, a value greater than 1 leads to a greater update weight for negative part, and a value less than 1 indicates a greater update weight for positive part. Through hyperparameter search, we found that the model achieved the highest mAP performance when A=1.05 for the current training settings.~\cref{fig:4ABL-study-fig}(d) shows how the position of $P_{neg}$ changes as training progresses during the preliminary training process, and it can be observed a tendency for the value of $P_{neg}$ to become progressively smaller, implying the adjustment of our dynamic loss function.

In the sampling unbiased processing, the hyperparameter values of $upper$ and $threshold$ can be used to adjust the mitigation of weight for hard samples. Usually, the higher the threshold value means the higher the magnitude of alleviation, and the higher the upper value means the higher the number of samples involved in mitigation. The influence of $upper$ and $threshold$ on performance is shown in~\cref{fig:4ABL-study-fig}(b). In addition, for models that can achieve similar performance in preliminary training, we found that selecting models that trained at a lower $threshold$ and a lower $upper$ conditions as possible could lead to larger performance improvement in subsequent annealing training. 

\textbf{Supervising network.} Original HNS tends to extract samples that are difficult to be identified by the current model as the number of training epochs increases. As the difficulty of the samples increases, the proportion of triplet tuples that are worth learning in the samples gradually becomes smaller, resulting in a better fit of the model to extremely difficult samples and a decrease in performance in real-world-like test scenarios.~\cref{fig:4ABL-study-fig}(c) shows that as the number of training epochs increases, the mAP performance of the model on Hpatches full split tends to increase and then decrease. Compared with the previous one, the introduction of the supervising network alleviates this problem in training and provides a more relaxed selection interval for the optimal model, without worrying about the timing of early stopping. At the same time,~\cref{fig:4ABL-study-fig}(c) shows the mean and variance of the performance using the balance loss (no unbiased processing) with 4 different random seeds trained after 22 epochs under the same settings, which has a larger variance compared with the balance loss. Thus, the supervising network can also help the supervised network to improve performance stability.

In addition, the selection of the hyperparameters is related to the performance and training settings of the supervising network used. In addition to using the supervised network itself as the supervising network (Self-TNet), we also choose a larger network trained on the Liberty dataset, which has twice the number of channels per layer compared to the student net and therefore outputs a 256-dimensional vector. The experimental result is shown as Pre-TNet. In terms of specific training settings, we observed that using the pre-trained network as the supervising network, which does not have the best performance, can guide the supervised network to achieve the highest mAP. Presumably, it is related to the difference in fitting ability caused by the difference in the number of parameters between the supervising network and the supervised network \cite{gou2021distillation-survey}. Finally, we choose the network trained by balance loss with a fixed $\beta$ value and mining negative number equals 2 as the supervising network to guide the supervised network. Although the trained supervising network cannot achieve the optimal performance, it is beneficial for the supervised network to receive more information about the overall data distribution from the supervising network in the guidance.
%When negative number=2, the network will focus more on the overall sample fit and less on the hard negative samples. 

\textbf{Annealing training. }The main hyperparameters in AT are the step sizes of $bs$ and $thr$. In this paper, we use $bs$ decreasing by 128 at a time and $thr$ increasing by 0.05 at a time as a general adjustment strategy. In general, smaller step sizes mean smoother transitions between the two training phases and usually better performance, but they also increase the computational cost. Here we balance the effects of both to choose the hyperparameters.

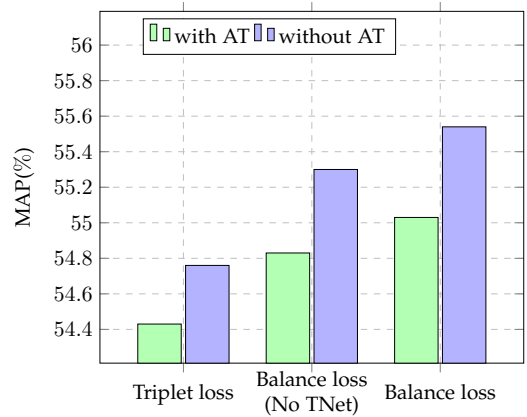
\begin{figure}[h]
    \centering
    \begin{minipage}[c]{0.8\linewidth}
    \centering
%%%%%%%%%%%%%%%%%%%%%%%%%%%
% \begin{figure}  \scriptsize
	%\begin{figure}
	% \centering
 % \subfigure{
\resizebox{\linewidth}{!}{
    \begin{tikzpicture}
\begin{axis}[
	  %x tick label style={/pgf/number format/1000 sep=},
        xtick={1, 2,  3 },
        xmin=0.5,xmax=3.5,
        xticklabels={Triplet loss, \makecell[c]{Balance loss\\(No TNet)}, Balance loss},
        xticklabel style={text height=2ex},
        ymin=54.3, ymax=56.1,
	ytick={54.4, 54.6, 54.8, 55.0, 55.2, 55.4, 55.6, 55.8,  56.0},
	ylabel=MAP($\%$),
	enlargelimits=0.05,
	legend style={at={(0.4,0.98)},
	anchor=north,legend columns=-1},
	% ybar interval=1.2,
        ybar, bar width=20pt,
        xmajorgrids=true,
	ymajorgrids=true,
	grid style=dashed,
]
\addplot [fill=green!30]
	coordinates {(1,54.43) (2,54.83) (3,55.03)
		  };
\addplot [fill=blue!30]
        coordinates {(1,54.76) (2,55.30) (3,55.54)
		 };
\legend{with AT, without AT}

\end{axis}
\end{tikzpicture}

   
 }
%%%%%%%%%%%%%%%%%%%%%%%%%%%
    \end{minipage}
    \caption{The performance of the original loss function before and after annealing training and the performance of balance loss before and after annealing training were compared} \label{fig:AN_comp}
\end{figure}
AT has a similar effect on the performance improvements shown in~\cref{fig:AN_comp}, as long as the model has previously undergone the HNS process.
In addition, we can see from~\cref{fig:AN_comp} that the model trained with balance loss can get more performance improvement in the annealing training. 
This can be attributed to the fact that the balance loss can handle the difficulty of the training samples by assigning more appropriate adaptive weights than the triplet loss, thus helping to learn more abstract laws from the samples. This proves that our two improvements can collaborate to provide the best performance. 
%For example, the model obtained by using balance loss as HNS strategy(ST in~\cref{fig:AN_comp}) can get more performance improvement than traditional triplet loss (original in~\cref{fig:AN_comp}). 
%Considering that 2D images are low-density information, proper hard negative mining can force models to discover abstract laws from more difficult triples [ ]. These higher-level laws can guide the model to identify relatively simple samples when the model goes through annealing training.
% \subsection{Impact of Hyperparameters}

\section{Conclusion}
In this paper, we investigated how to obtain and utilize high quality negative samples in the widely used hard negative sampling strategy in metric learning, trying to balance the difficulty of the HNS extraction samples to provide the learning samples which are most suitable for the training of the current model. The proposed balance loss combines the self-supervised method into a specific dynamic gradient modulation strategy to achieve fine-grained gradient modulation for different difficulty samples. The proposed annealing training provides data sources with different difficulty distributions for the loss function, which can alleviate the overfitting of hard samples in the training process and help the model to learn the general laws from the data. Our improved local descriptors show superiority on a variety of tasks and datasets.

% or


%%%%%%%%%%%%%%%%%%%%%%%%%%%%%%%%%%%%%%%%%%%
% \appendices

% % you can choose not to have a title for an appendix
% % if you want by leaving the argument blank
% \section{}

% % use section* for acknowledgment
% \ifCLASSOPTIONcompsoc
%   % The Computer Society usually uses the plural form
%   \section*{Acknowledgments}
% \else
%   % regular IEEE prefers the singular form
% \fi


\ifCLASSOPTIONcaptionsoff
  \newpage
\fi

\bibliographystyle{IEEEtran}
\bibliography{IEEEabrv, ref.bib}
\end{document}